# Application of image-to-image translation in improving pedestrian dete[1]ction

**Devarsh Patel[1*], Sarthak Patel[2], Megh Patel[3]**

[1]Department of Data Science, Indian Institute of Science Education and Research, Pune, India
[2]Department of Computer Science, Vellore Institute of Technology, Bhopal, India
[3]Department of Computer Science, Birla Institute of Technology and Science Pilani – KK Birla Goa Campus, Goa, India

*Corresponding author at
Email: devarsh.patel@students.iiserpune.ac.in

## ABSTRACT

The lack of effective target regions makes it difficult to perform several visual functions in low intensity light, including pedestrian recognition, and image-to-image translation. In this situation, with the accumulation of high-quality information by the combined use of infrared and visible images it is possible to detect pedestrians even in low light. In this study we are going to use advanced deep learning models like pix2pixGAN and YOLOv7 on LLVIP dataset, containing visible-infrared image pairs for low light vision. This dataset contains 33672 images and most of the images were captured in dark scenes, tightly synchronized with time and location.

## 1. Introduction

Application of visual tasks are challenging in low light scenarios due to loss of information. However, by providing additional information, we can improve the performance of vision models. Infrared imaging can help to tackle this issue. Unlike visible images which retain details and texture information of objects, infrared images provide thermal information of objects and highlight targets like pedestrians. But without the availability of specialized cameras, it is difficult to obtain such images [1]. The major problem with infrared cameras is that they are quite expensive as compared to visible cameras which are easily accessible and can be even found on phones. Apart from that, infrared cameras use high-tech technologies to capture images even in the low light. As a result, they are expensive and not everybody can afford it. Visible cameras or day-to-day use cameras don't have good sensors and optical powers to capture high resolution images at night and as a result the image is quite blurry or contains significant grains or noises which makes them almost useless for security purposes [2]. This also makes it impossible for the object detection algorithms to work effectively for detecting pedestrians, as a result we don't get satisfactory results. So, In this model we are going to convert visible images to infrared images with the help of Generative Adversarial Networks (GANs). That converted image can be used for effective object detection as well as video surveillance. By this approach, we can easily solve the problem of detecting pedestrians in the low light areas.

For tackling this problem, we are going to use a visible-infrared paired for low-light vision (LLVIP) dataset [3]. The images in the dataset are taken with the help of a visible light camera and it has been ensured that the images are consistent with space and time. The dataset is helpful for low-light pedestrian recognition since it includes many distinct pedestrians in low-light situations. Image labeling is one of the challenges in this detection work since human eyes can barely differentiate pedestrians, much less precisely mark the bounding boxes.

Besides LLVIP, we have also explored other related datasets like INO Videos Analytics Dataset, TNO

[1] Unpublished working draft. Not for distribution.

Image dataset and OTCBVS Benchmark Dataset before finalizing LLVIP dataset. Among all the dataset we explored, we found LLVIP as suitable for our research. INO (National Optics Institute of Canada) [4] Video Analytics dataset contained various infrared and visible videos and had very rich information and environment but not many images contained the pedestrians in low light background. TNO [5] multiband image data collection was more of a military purpose because most of the images consisted of military scenes recorded in multiband cameras. But apart from that, TNO consists of only 261 images of visible and infrared images along with a queue of similar types of images. OTCBVS [6] is a public use dataset for evaluating computer vision algorithms initiated by Dr Riad I Hammoud in 2004. It is a visible infrared paired dataset and most of the images were captured at a crossroad of the Ohio state university. The images are rich in information along with the large number of images containing pedestrians but all of the images were clicked in daytime. Consequently, due to images clicked in the daytime, it's very easy to detect pedestrians because the images by default were very clear. So, in this case we wouldn't have been able to use the infrared images. As a result, LLVIP is the balanced dataset of all the sample datasets mentioned above.

### 1.1. The objective of the paper

The main objectives of this paper are 3-fold:

1. To propose a model that can detect pedestrians in the low light region with the help of image-to-image translation.
2. To propose object detection with the help of advanced deep learning models like YOLOv7 and Pyramid pix2pixGAN.
3. Extensive review of proposed deep learning models using various performance measures

### 1.2. Proposed novel work

Getting infrared images requires expensive camera setup and hence makes it infeasible to upgrade or replace existing surveillance infrastructure. However, as demonstrated in previous works [3], it is possible to use GANs to perform image-to-image translation on visible images to get a prediction for corresponding infrared image. We hypothesize that doing object detection on translated infrared images will improve performance of pedestrian detection tasks without using any special equipment.

Our aim is to detect pedestrians from camera images with relative high accuracy using object detection models on infrared images obtained through image translation. We use LLVIP dataset to train pyramid pix2pixGAN [16] which is shown to be state-of-the-art on this dataset. Then using this trained model, we convert a visible image to corresponding infrared image and train a YOLOv7 model on these images. We evaluate performance of object detection using this ensemble technique and compare it with models trained on just visible images.

### 1.3. Paper Organization

The remainder of the paper is organized as follows; Section 2 provides a summary of existing similar efforts. Section 3 goes through the dataset and its characteristics in greater detail. Section 4 describes the deep learning models that were used. Section 5 describes our training methodology. It is concerned with data preprocessing and dataset preparation prior to executing the deep learning model. Section 6 contains the efficiency measurements, methods, and graphical analyses. Section 7 finishes with a review of our findings and projections for the breadth of future improvements.

## 2. Related work

Several studies on detecting pedestrians in low light conditions have been conducted so far using various machine learning as well deep learning models. Some of them are listed below:

1. Vedant et al. [7] demonstrated detecting pedestrians in low light conditions using the traditional computer vision and deep learning techniques, like using strength of signals for a depth sensing camera as well as robust principal component analysis (RPCA) for dimensionality reduction. They used the masking techniques to modify Red Green Blue (RGB) images to boost the distinction between foreground and background images. They implemented their deep learning models 'CrowdDet and CenterNet' on 'Oyla Low-Light Pedestrian Benchmark' (OLPB) dataset.

2. Srinivas et al. [8] used a multi-modal knowledge distillation technique to detect pedestrians from RGB images. They are using a residual neural network (ResNet) which takes images as RGB image data and extracts thermal-like features to detect pedestrians. For their research they used the KAIST dataset and achieved a 52.81% miss rate (lower the better) for their ResNet model.

3. Congqing et al. [9] suggested a way to detect near-surface pedestrians for unmanned aerial vehicles (UAVs) using computer vision. Most of the UAVs are well equipped with the infrared sensors. This research directly uses infrared images and applies You Only Look Once Version 3 (YOLOv3) for object detection. They achieved the accuracy (P) of 0.804 and F1 score of 0.859.

4. Yonglong et al. [10] put forward a mechanism to detect pedestrians using DeepParts and Convolutional Neural Network (ConvNet) on caltech benchmark dataset. Pedestrian detection works very well in caltech benchmark dataset with the help of deep neural network techniques because the caltech dataset contains video sequences recorded in a well-lit environment; when faced with inputs recorded in dimly-lit environments, the aforementioned algorithms frequently perform inadequately.

5. Yi jin et al. [11] approach the problem of detection of street pedestrians in low light conditions with the help of a super-resolution detection (SRD) network. Their model is based on a playground dataset (PG) which contains the 5,752 combined images of pedestrians taken in day and night with 31,041 annotations. Their strategy was to enhance the low-resolution images to high quality images for efficient object detection. Enhanced images went through Recurrent Convolutional Neural Network (R-CNN) to locate obstructed pedestrians.

Contrarily, our suggested algorithm converts the low light visible images to infrared images with the help of pyramid pix2pixGAN - a Generative Adversarial Network and after that applies the latest and one of the most advanced object detection algorithms - YOLOv7 [12] to detect pedestrians much effectively with the better accuracy and faster results.

## 3. About dataset used

We have used LLVIP [3], a dataset containing paired visible-infrared images taken in low light scenarios, suitable for training pedestrian detection models and transforming visible light image to infrared image. The dataset also provides annotations detailing the bounding box around each person present in a particular image. There are in total 15488 pairs of visible-infrared images taken from 26 different locations. Resolution of each image is 1280-pixel width by 1024-pixel height.

### 3.1. How the images were captured and annotated

The authors of the dataset used HIKVISION DS-2TD8166BJZFY-75H2F/V2, a binocular camera that has an infrared camera (wavelength 814 um) and a visible light camera. The pictures taken between 18:00:00 and 20:00:00, contain street shots from various spots and include lots of bicycles and pedestrians. To ensure that visible-infrared image pairs had the exact same area of view and image size, the images were semi-manually cropped to ensure temporal and spatial synchronization. For annotation, the authors have applied a two-fold approach, first they have identified pedestrians on infrared photos where they are clearly visible, then the annotations have been copied straight to the visible image due to the alignment of the infrared and visible images.

### 3.2. Some problems with the dataset

The pedestrians in the majority of the dataset are of a medium size as the photos were taken from a medium distance. As a result, the investigation of long-distance small-target pedestrian identification is not appropriate for this dataset. Another problem is that all of the images were taken from the perspective of a stationary surveillance camera and thus are not appropriate for training pedestrian detection for self-driving cars

## 4. About models

In our research we have used pyramid pix2pixGAN to generate thermal images of the visible images and YOLOv7 for the efficient object detection.

### 4.1. Pyramid pix2pixGAN (Generative Adversarial Networks)

Generative Adversarial Networks (GANs) [26] are the neural networks which help us to generate new images or data from the base image. First, by imitating existing pictures, Generative Adversarial Networks (GANs) is an image-to-image translation neural network that enables us to create new and more precise sets of images from the current ones [13]. The Generative model creates new pictures from the basic images, while the Discriminator, as the name implies, determines if an image is real or counterfeit. A form of GAN called Pix2PixGAN [22] aids in both text to picture and image to image translation. But in contrast to the classic GAN model, which classifies images using a deep convolutional neural network, The Pix2Pix model uses a conditional GAN (CGAN). This deep convolutional neural network is designed to recognise specific areas of the input picture rather than categorizing the entire input image as true or false [14-15].

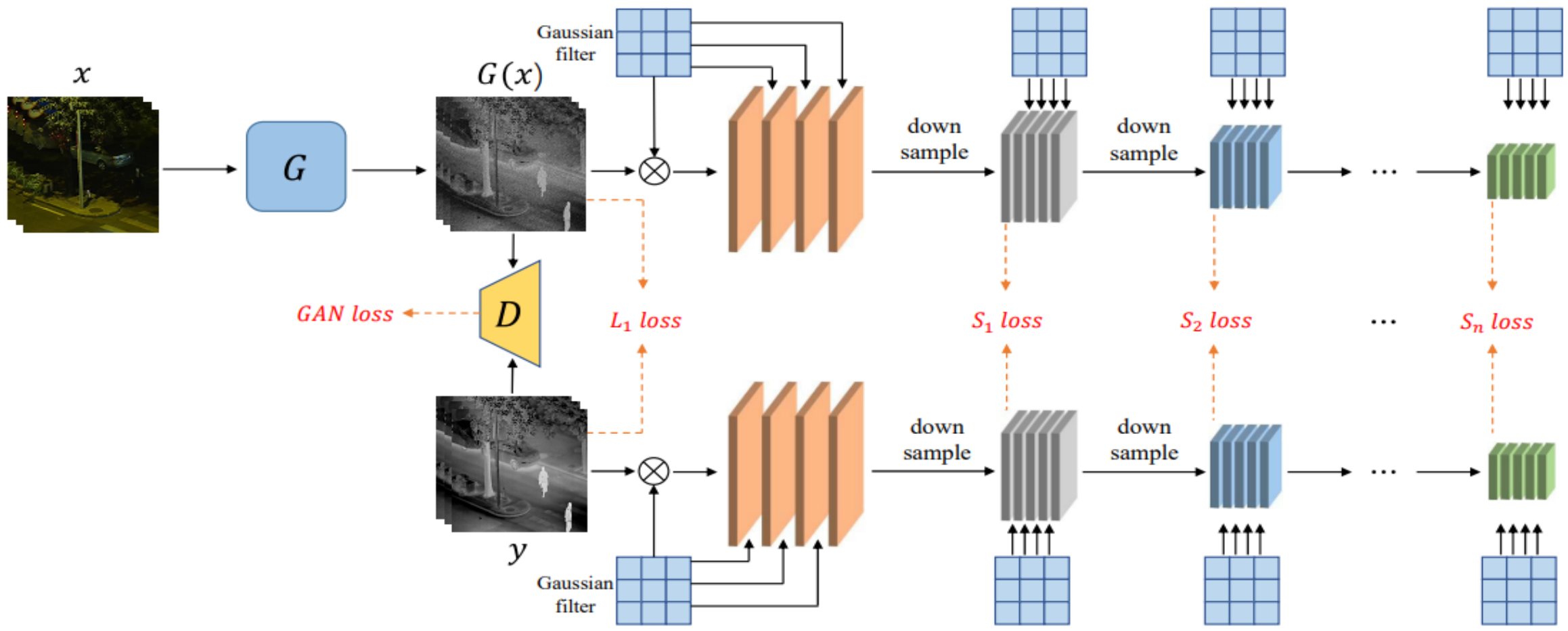

**Figure 1.** Pyramid pix2pixGAN architecture

But due to the difference between images of two domains, it is difficult for the normal pix2pix algorithm to work as the positions in the image are not able to obtain the pixel-to-pixel level alignment. As a result, we came to a conclusion to use pyramid pix2pixGAN. The pyramid architecture can be seen in **Figure 1**. The Loss factor L1 of the pix2pix algorithm computes the disparity between the original image(real) and generated image (fake), which is overly constrained in the generated image. To weaken the constraints, we are going to apply scale transformation on real and fake images. Firstly, we are going to use a Gaussian filter to smooth the images, but as a result images are going to be a bit blurred, and to reduce that resolution, we use downsampling to get rid of unnecessary pixels [16]. For every single octave (resolution level), the above steps are repeated multiple times to attain efficient scale transformation. Each octave of our pix2pix pyramid comprises five layers and four Gaussian blurings, with the initial layer of each octave being created by downsampling the final picture of the octave before it. To determine the loss in our Gaussian Pyramid, we extricate the only first layer of images for each octave. For each layer (scale), loss can be denoted with Si:

$$S_i = \mathbb{E}_{x,y,z}\left[\left|\left|F_i(y) - F_i\big(G(x,z)\big)\right|\right|_1\right]$$

Where G represents Generator and Fi is for Gaussian Filtering. x denotes our input image. y and z signify real image and noise, respectively.

### 4.2. You Only Look Only Once (YOLOv7)

YOLOv7 is a neural network implemented algorithm which provides faster and better accuracy for object detection. YOLOv7 divides the image into various grids of fixed size. Every grid is responsible for detecting objects enveloped in them. The bounding boxes and confidence scores for each box are predicted in each grid cell. These confidence scores demonstrate the model's level of assurance that the box contains an object as well as how accurately it affirms that the box contains that particular object. If there is no object within the box, then the confidence score is given as zero [17]. This approach significantly reduces computation since cells from the image simultaneously maintain both recognition and detection, but it generates a lot of duplicate predictions because multiple cells may predict the same object but with distinct bounding boxes. To tackle this issue YOLOv7 uses the concept of non-maximal suppression. In this technique, the bounding boxes with lower probability scores are suppressed or not considered. To do this, YOLOv7 selects a bounding box with the highest probability score. After this, the bounding boxes having the biggest Intersection over Union (IoU) with the current high probability bounding box are then suppressed. The steps are repeated until we get the final bounding box with the accurate object detection [18-19].

YOLOv7 algorithm is a composition of several bag-of-freebies methods which can vastly enhance accuracy of the detection using less computation power. The method outputs superior results than all other object detection models by using about 40% of the function parameters without compromising accuracy. The team of YOLOv7 has created very effective architecture (**Figure 2**). They extended the efficient layer aggregation network (E-ELAN) which employs expand, shuffle and merge number of attributes to recursively improve the network's learning capacity without altering the initial gradient; that is, it only modifies the block of computation without touching the transition layer [20].

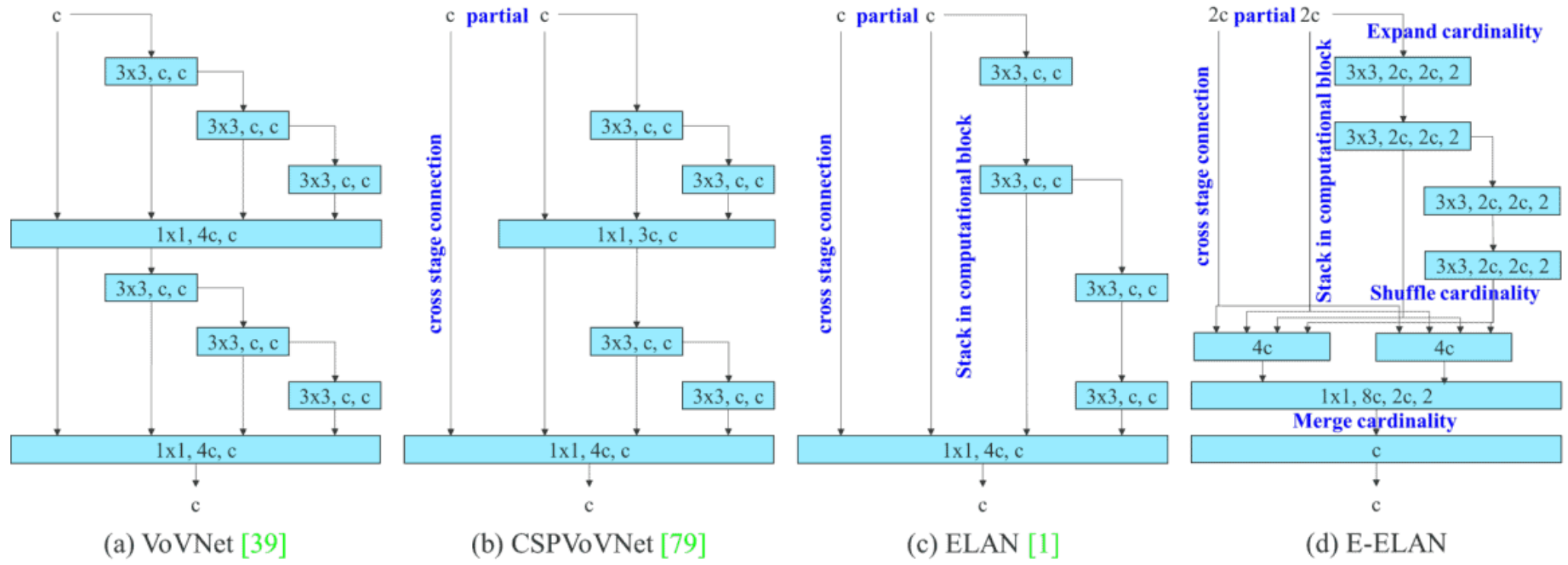

**Figure 2.** YOLOv7 architecture

## 5. Methodology

This section covers methods used to preprocess data to train pyramid pix2pixGAN and YOLOv7 model. We also discuss the experimental details that we conducted on aforementioned models.

### 5.1. Data preprocessing

Our ground truth image is an infrared map which we desire from our model to output by performing image-to-image translation on visible images. The corresponding infrared images are grayscale images thereby we set output channels to be 1 in pyramid pix2pixGAN. Visible images were loaded according to input size mentioned in Section 5.2.

### 5.2. Experimental setup

A system with 4 Tesla V100-SXM2-16GB GPUs and 64GB of vRAM was used to train the network with different models. Pytorch [21] library was used to implement the complete design.

We trained 2 models namely pyramid pix2pixGAN for image-to-image translation and YOLOv7 for object detection

#### 5.2.1 Image-to-image Translation

We used the same train test split from the LLVIP dataset. In total, there were 12025 images for training and 3463 for test. We first resize images to load size of 320×256, and then we scale width of image to 320 and crop image to 256×256 as data preprocessing step.

The model uses the baseline generator structure of resnet-9blocks and the structure of the discriminator is the basic PatchGAN as default. We train models with batch size set to 64. Adam optimized was used with an initial learning rate of 0.0002. Model was trained for 100 epochs with the initial learning rate 0.0002 and linearly decay learning rate to zero for next 100 epochs.

#### 5.2.2 Object detection

For generating training data for object detection, we did inference using trained pyramid pix2pixGAN model and translated all visible images in train and test dataset to make new train and test data for YOLOv7 to train on. We used annotations given in the LLVIP dataset as bounding box labels for the training model.

We used pretrained checkpoint YOLOv7 [12]. We finetune this model training data generated above. The models are trained for 150 epochs on a batch size of 16. We use SDS to optimize the model with an initial learning rate of 0.001 with weight decay of 0.0005 and momentum of 0.825.

## 6. Results and Discussion

In this section, we are going to discuss the results of the image-to-image translation using pyramid pix2pixGAN and pedestrian detection using the latest

YOLOv7 model in great detail statistically as well as graphically.

## 6.1. Image Translation

To detect pedestrians efficiently in challenging conditions, like with low visibility it is beneficial to convert the visible images to infrared images. Because in the low light conditions the pedestrians somewhat blend with the background and the model finds it difficult to distinguish the pedestrians with objects, as a result we need to translate the image. In order to get things done, we need to translate the images from visible to infrared with the help of GANs or segmentation models. However, In previous works on this domain it has been observed that the normal pix2pixGAN doesn't perform good conversion as some of translated images were blurry or messy. For some images the contours of pedestrians and cars were not clear with many artifacts on the image. As a result, for our research we have used pyramid pix2pixGAN which is somewhat similar to pix2pixGAN but more versatile because normal pix2pixGAN was not able to obtain the pixel-to-pixel level alignment for the images for the dataset.

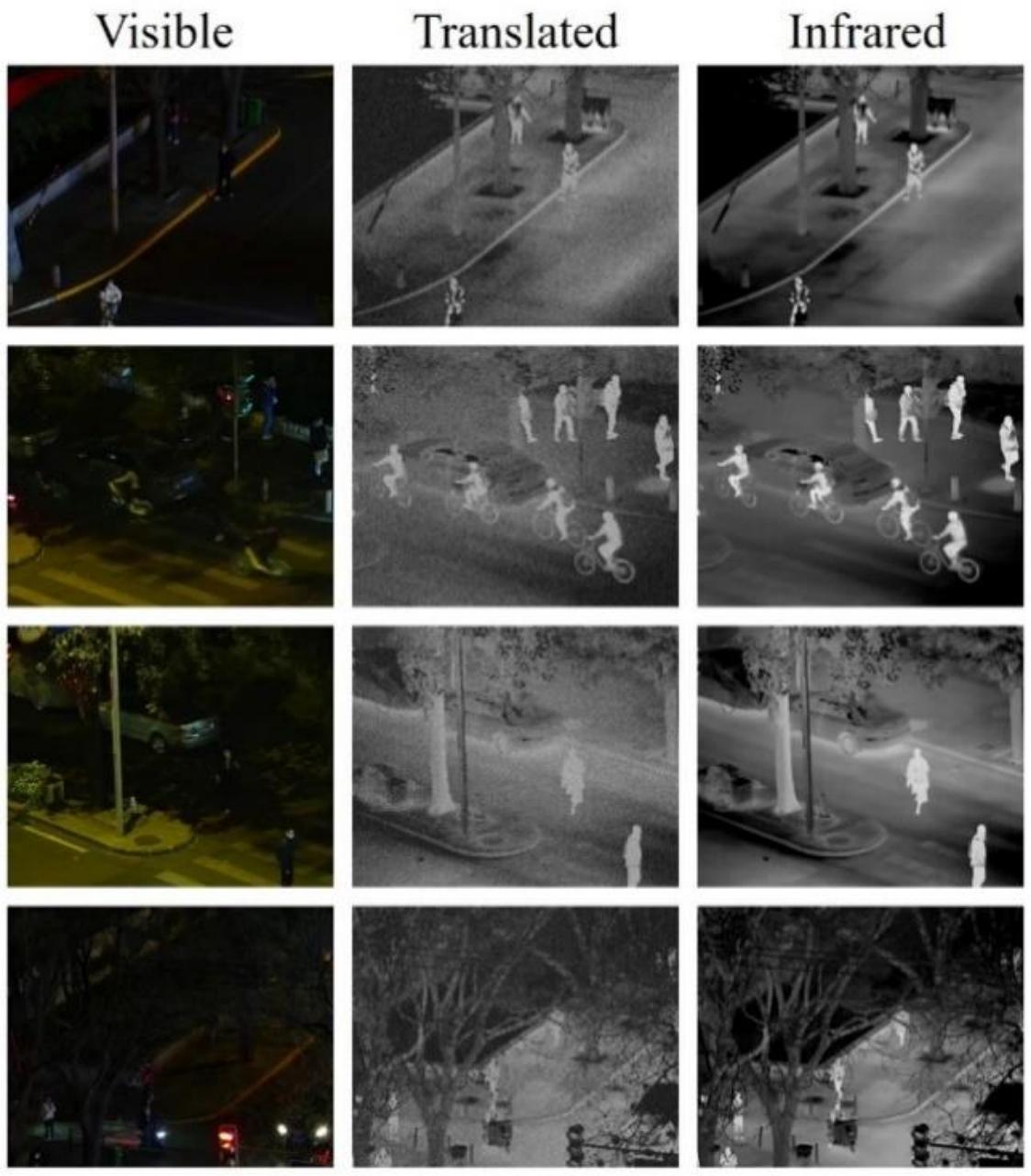

**Figure 3.** Illustration showing the results of visible to infrared image translation for varying lighting scenarios. Left to right, in order: Actual visible images, Translated infrared images, Actual infrared images

**Figure 3** qualitatively compares the visible images with translated infrared and original infrared. In certain visible images, it was even challenging for humans to see pedestrians, but turning such images into infrared ones made it much easier to see pedestrians in low light.

Pyramid pix2pixGAN provided satisfactory results for the conversion. The translated images are almost similar to the original infrared images. The contours of the generated images are preserved as well as images are not messy or blurry but the downside of the generated infrared images is that they are a bit diminished in quality in comparison with the ground truth images. However, the translated images were successful in serving our purpose.

| Metric | MSE | PSNR | SSIM |
|---|---|---|---|
| Result | 0.1734 | 12.0739 | 0.2591 |

**Table 1**. Image translation results

The quantitative performance of Pyramid pix2pixGAN using different metrics can be understood from **Table 1**. MSE is used to calculate the combined square error of original and compressed images. Peak signal-to-noise ratio (PSNR) is a metric for the ratio of a signal's greatest permissible value (power) to the power of distorted noise that impairs the accuracy of its representation. A perceptual metric called the Structural Similarity Index (SSIM) measures how much image quality is lost during processing like data compression or during data transfer.

## 6.2. Object Detection

Now after translating images from visible to infrared, we detect pedestrians using the YOLOv7 model on both visible and infrared images and compare the performance on both the instances. The qualitative comparison of object detection on both types of images has been displayed in **Figure 4**. The **Figure 4** contains images in different light conditions. Since the image in the first row was taken in good light conditions, YOLOv7 detected all pedestrians accurately in both visible and translated images. The Image in the third row was clicked in slightly poor light conditions as compared to the image in row 1. This time YOLOv7 could only detect one pedestrian instead of 2 in visible light while it detected all in the infrared image. The images in row 2 and row 4 are taken in very dim light conditions. In both the images YOLO failed to detect all the pedestrians in visible conditions while it detected all the pedestrians with a very high confidence score as compared to the former.

The infrared image contours pedestrians and has a greater effect on the detection job, which not only demonstrates the need for infrared images but also signifies that the pedestrian detection algorithm's performance in low-light situations is not fully efficient.

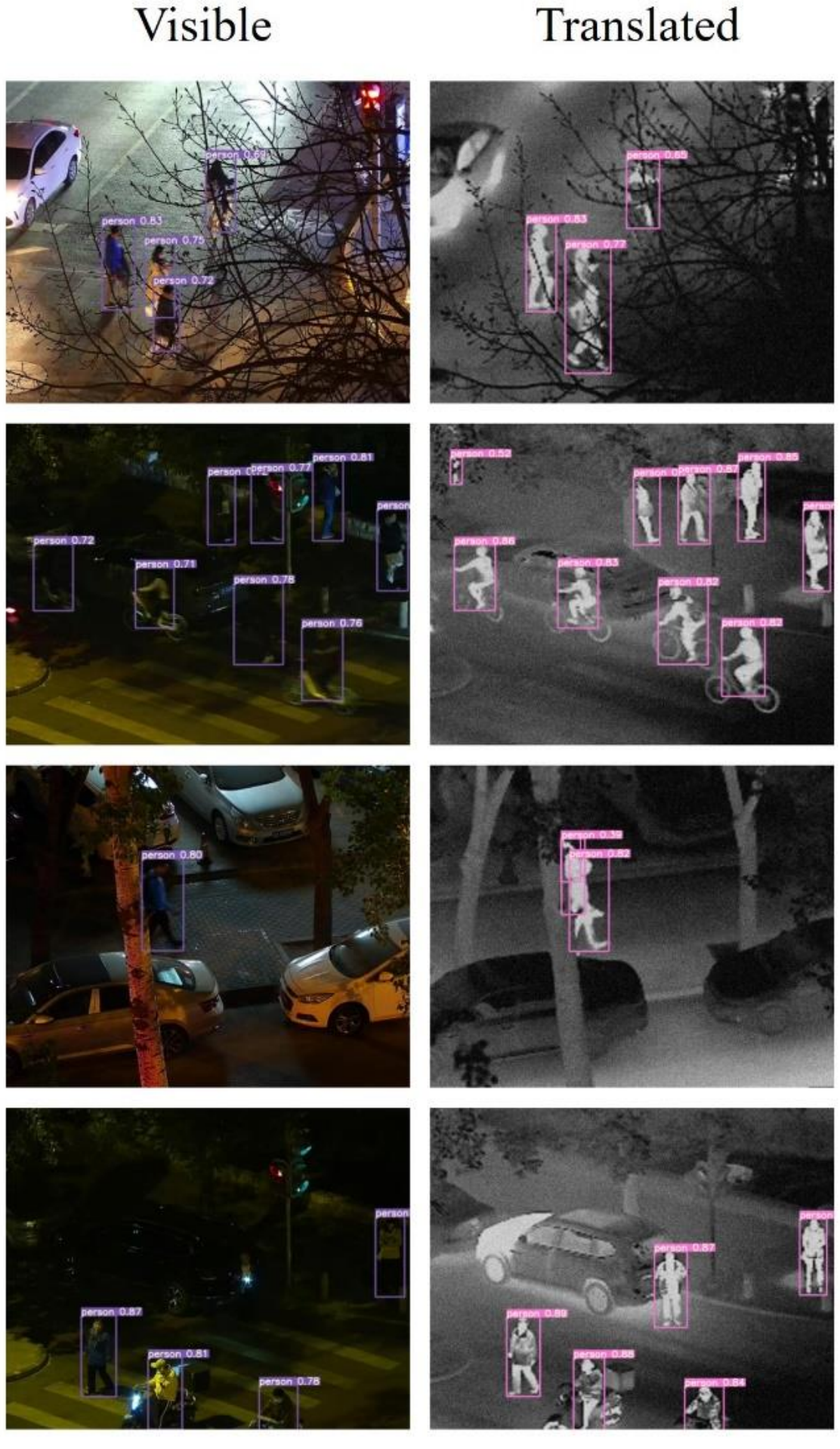

**Figure 4.** Illustration showing the results of pedestrian detection in both translated infrared images and visible images for varying lighting scenarios. Left to right, in order: Detection in visible images, Detection in translated infrared images

For quantitatively analyzing our models, we are going to use assessment metrics like Mean Average Precision (mAP@0.5 and mAP@0.5:0.95), Precision and Recall graphically. These are explained below

1. The proportion of accurately categorized positive samples (True Positive) to the total number of positively classified samples (either correctly or incorrectly, True Positive + False Positive) is known as precision. Precision enables us to see the machine learning model's dependability in correctly identifying the sample data as positive [23].

2. The recall is calculated by taking ration of True Positive to all Positive samples (True Positive + False Negative). Recall gauges how well the model can identify positive samples [24].

3. The mAP@0.5 calculates a score by comparing the detected box to the ground-truth bounding box at IoU threshold of 0.5. The sample is categorized as true positive if the overlap between the predicted bounding box and the ground truth bounding box is more than 0.5, otherwise as false positive. The model's detections are more precise the higher the score [25].

4. mAP@0.5:0.95 refers to the average mAP over various IoU thresholds, from 0.5 to 0.95, in steps of 0.05.

The quantitative comparison of detection in visible and infrared conditions has been displayed in the combined line graphs for different metrics in fig 5. The model has been run for 100 epochs and the metrics results for each epoch has been plotted for both types of images. It is clearly evident that object detection on the infrared images (blue line) outclassed visible images (orange line) for detecting pedestrians in all the evaluation metrics.

## 7. Conclusion

In this paper, we propose to detect pedestrians from visible camera images taken in low light conditions with relative high accuracy using object detection models on infrared images obtained through image translation. We hypothesize that doing object detection on translated infrared images will improve performance of pedestrian detection tasks without using any special expensive equipment for infrared imaging. We evaluate performance of object detection using this ensemble technique and compare it with models trained on just visible images and find that our technique outperforms visible image models in all aspects especially in very low light conditions.

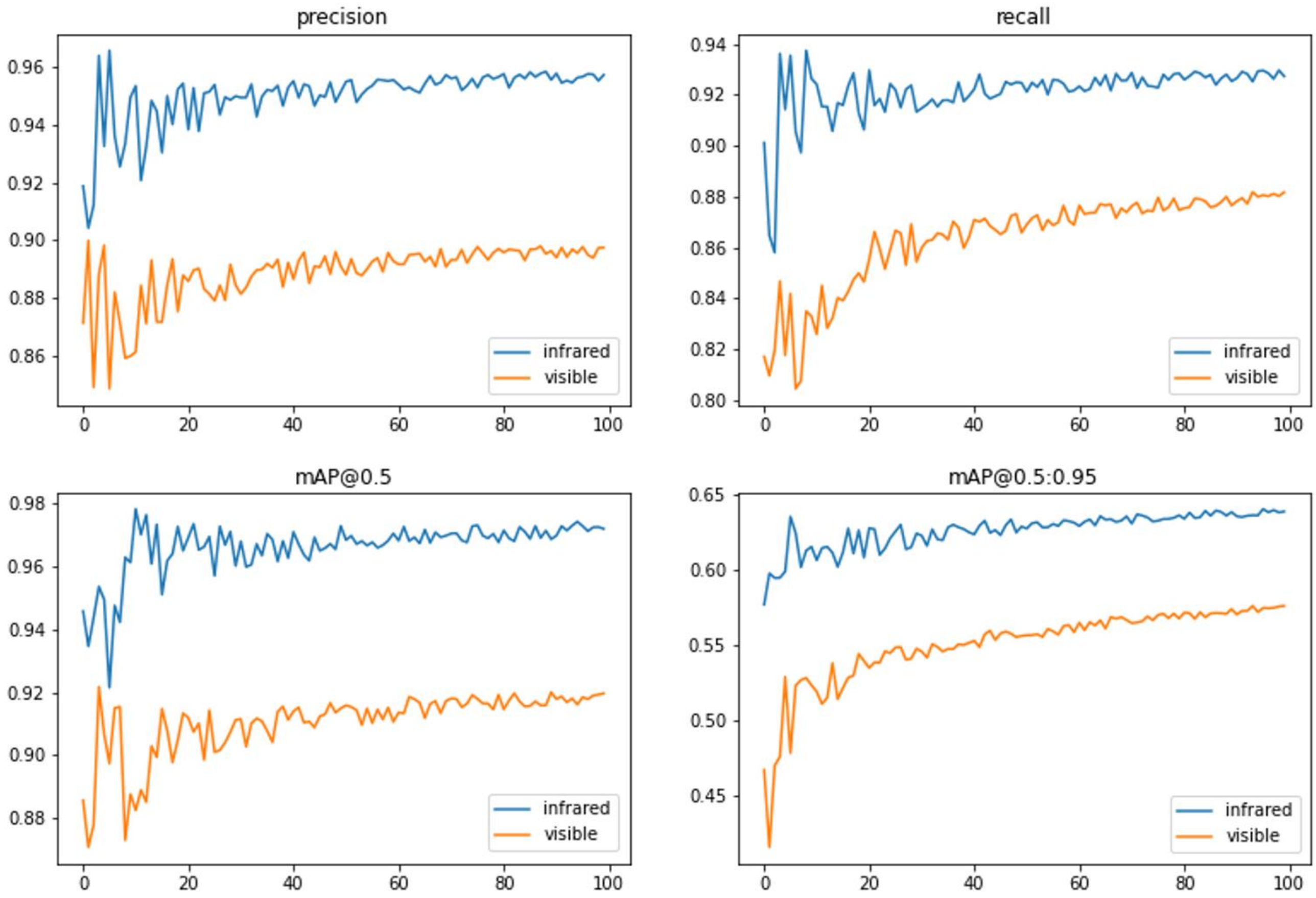

***Figure 5.*** *Object detection result comparison between translated infrared (blue) and actual visible (orange) images*